# Toward Smart Power Grids: Communication Network Design for Power Grids Synchronization

## 12-E-SMG-1366

Hojjat Salehinejad[1], Farhad Pouladi[2], and Siamak Talebi[1,3]

[1]Electrical Engineering Department, Shahid Bahonar University of Kerman, Kerman, Iran
[2]Scientific-Applied Faculty of Post and Telecommunications, Tehran, Iran
[3]Advanced Communications Research Institute, Sharif University of Technology, Tehran,Iran
h.salehi@eng.uk.ac.ir; pouladi@ictfaculty.ir; siamak.talebi@uk.ac.ir

**ABSTRACT**

In smart power grids, keeping the synchronicity of generators and the corresponding controls is of great importance. To do so, a simple model is employed in terms of swing equation to represent the interactions among dynamics of generators and feedback control. In case of having a communication network available, the control can be done based on the transmitted measurements by the communication network. The stability of system is denoted by the largest eigenvalue of the weighted sum of the Laplacian matrices of the communication infrastructure and power network. In this work, we use graph theory to model the communication network as a graph problem. Then, Ant Colony System (ACS) is employed for optimum design of above graph for synchronization of power grids. Performance evaluation of the proposed method for the 39-bus New England power system versus methods such as exhaustive search and Rayleigh quotient approximation indicates feasibility and effectiveness of our method for even large scale smart power grids.

## I. INTRODUCTION

Synchronization in power grids is one of the critical issues in system stability. A network in synchrony will be able to deliver stable electric power after a disturbance and avoid large scale black out. The synchronization for the power networks can be provided with or without assist of communication infrastructures. In case of none-communication network, physical links like transmission lines are used to couple generators. In this case, the synchronicity is determined using the 'swing equation' and the largest eigenvalue of the Laplacian matrix of the power network [1]. On the other side, in case of having an available communication infrastructure, the measurements of generators are sent to other generators for feedback control. In this case, the synchronicity is determined by the largest eigenvalue of the weighted sum of the Laplacian matrices of the power network and the communication network [1]. In [1], the communication network is modeled using 'swing equation' and Laplacian matrices of the power and communication networks. In [2], the reduce admittance matrix impact, on the synchronization of power networks is studied. In [3] and [4], a continuum model is proposed for the power network and partial differential equations are employed for describing the dynamics. In [1], the derivation of the generator couple is similar to those of [3] and [4], while impact of network topology is also considered. The main difference between [1] and existing literature is that it has considered communication network in its model.

In graph models, the problem of finding a route with specific characteristics is considered as a NP-hard problem, which needs considering all the possible routes [5]. Metaheuristic algorithms are of great interest for solving large scale problems and particularly the NP-hard ones in academia and industry [5, 6]. The crucial benefits of mentioned nondeterministic structures are their less complexity, low computational load, and few time usage for solving complex problems, which may have no derivatives, in a short period [5,6,8,9].

This paper investigates a new method initiated on Ant Colony System (ACS) for optimum design and planning of communication infrastructure for power grids synchronization using the communication network graph in [1]. In this method, it is attempted to study how to design the topology of the communication network such that the power network can be efficiently synchronized.

The rest of the paper is organized as follows. In Section II a short survey on ACS is introduced. Next, a model for power grid is presented in section III. Then, the proposed method is discussed in section IV. Later, performance of proposed method is evaluated in section V. At the end, the paper is concluded by mentioning some research challenges in Section VI.

## II. ANT COLONY SYSTEM

In natural world, ants find shortest path from a food source to their nest by using information of pheromone liquid that they discharge on the journey path. The ACS endeavors to work base on some artificial characteristics, which are adapted from real ants with the addition of visibility, discrete time, and memory. This computational approach to problem solving has many applications in complex problems such as in wireless communications [8] and intelligent systems [9]. According to [5] and [9], a typical ACS algorithm is consisted of the following steps, interested readers are referred to [9] for more details:

*Problem Graph Depiction:* The artificial ants are mostly employed for solving discrete problems in discrete environments. This can result to considering the discrete problems as graphs with a set of *N* nodes and *R* routes.

*Ants Distribution Initializing:* In order to move ants through the graph, some ants must first be placed on a set of random origin nodes. In most applications, trial and errors as well as nodes density in the region define the number of ants.

*Ants Probability Distribution Rule:* When an ant wants to move from a node to another, a node transition rule is necessary. The transition probability of ant *k* from node *i* to node *j* is given by

$$p_{ij}^k = \begin{cases} \dfrac{(\tau_{ij})^\alpha (\eta_{ij})^\beta}{\sum_{h \notin tabu_k} (\tau_{ih})^\alpha (\eta_{ih})^\beta} & j \notin tabu_k \\ 0 & otherwise \end{cases} \quad (1)$$

where $\tau_{ij}$ and $\eta_{ij}$ are the pheromone intensity and the cost of direct route between nodes *i* and *j*, respectively; relative importance of $\tau_{ij}$ and $\eta_{ij}$ are controlled by parameters $\alpha$ and $\beta$ respectively; $tabu_k$ is set of unavailable edges for ant *k*.

*Update Global Trail:* When every ant has assembled a solution at the end of each cycle, the intensity of pheromone is updated by a pheromone trail updating rule, which is given by

$$\tau_{ij}^{new} = (1-\rho)\tau_{ij}^{old} + \sum_{k=1}^{m} \Delta \tau_{ij}^k \quad (2)$$

where $0 < \rho < 1$ is a constant parameter named pheromone evaporation. The amount of pheromone laid on the route between nodes *i* and *j* by ant *k* is

$$\Delta \tau_{ij}^k = \begin{cases} \dfrac{Q}{f_k} & \text{if route } (i,j) \text{ is traversed by the } k^{th} \text{ ant (at the current cycle)} \\ 0 & \text{otherwise} \end{cases} \quad (3)$$

where $Q$ is a constant parameter and $f_k$ is the cost value of the found solution by ant *k*.

*Stopping Procedure:* In typical algorithms, ACS procedure is completed by arriving to a predefined number of runs.

## III. SYSTEM MODEL

To plan an optimum communication network, first we need to model the communication synchronized power grid. To do so, first a power grid including the dynamics of generators is presented as a graph. Then, the synchronization with communication infrastructure is studied.

### A. Graph Modeled Power Network

A typical power grid is consisted of some geographical sites as well as transmission lines. To model such network as a graph $G$, we can consider each site as a node *A* and each transmission line as an edge *B* of graph $G < A, B >$. In this model, with respect to [3], each generator supplies a time-varying current and time-varying power, using a constant voltage. Each site is equipped with one synchronous generator and two adjacent generators *i* and *j* is denoted by $i \sim j$ [1].

Using the 'swing equation' denoted in [7], status of each generator can be described by the rotor rotation angle $\delta$ as

$$M\ddot{\delta} + D\dot{\delta} = P_m - P_e \quad (4)$$

where $M$ is the rotor inertia constant, $D$ is the mechanical damping constant, $P_m$ is the mechanical power, and $P_e$ is the electrical power defined as a function of the angle $\delta_i$ and neighboring nodes [1]. For the transmission line between two adjacent nodes $i \sim k$ in the network, the flowing current is given

$$I_{i,k} = \frac{E_i - E_k}{Z} \quad (5)$$

where $Z = R + jX$ is the impedance of the transmission line, $E_i$ and $E_k$ are the voltages of the generators $i \sim k$ and their magnitudes are considered constant as $E_i = Ve^{j\delta_i}$. By considering the Kirchoff's current law and defining $Y$ as the shunt admittance, the current conveying from the voltage source to the node is defined as

$$I_i = \sum_{i \sim k} I_{i,k} + YE_i = \frac{1}{Z}\sum_{i \sim k}(E_i - E_k) + YE_i \quad (6)$$

Therefore, the electric power $P_e^i$ is determined as

$$P_e^i(\{\delta_i\}_{i \sim j, i \ne j}) = \text{Re}[E_i I_i^*] = \frac{V^2}{|Z|^2}$$
$$\times \left[ R(N_i - \sum_{i \sim k}\cos(\delta_i - \delta_k)) - X\sum_{i \sim k}\sin(\delta_i - \delta_k) \right] \quad (7)$$
$$+ V^2 \text{Re}[Y] \approx -\frac{V^2 X}{|Z|^2}\sum_{i \sim k}(\delta_i - \delta_k) + V^2 \text{Re}[Y]$$

where $N_i$ is the number of neighbors of node *i* and $\delta_i - \delta_k$ is assumed too small to keep just the first order term in the Taylor's expansion. The dynamics of the power grid with respect to (4) and (7) is defined as

$$M\ddot{\delta}_i + D\dot{\delta}_i = P_m^i + \frac{V^2 X}{|Z|^2}\sum_{i \sim k}(\delta_i - \delta_k) - V^2 \text{Re}[Y] \quad (8)$$

This equation clearly illustrates that the dynamics of different nodes *i* and *k* are coupled by the difference of rotation angles $\delta_i$ and $\delta_k$ [1].

By letting $\xi_i = \delta_i - \delta_0$ for the mechanical power $P_m$ we have

$$P_m(\delta_i) = P_m(\xi_i) = V^2 \text{Re}[Y] + h\xi_i + o(\xi_i) \quad (9)$$

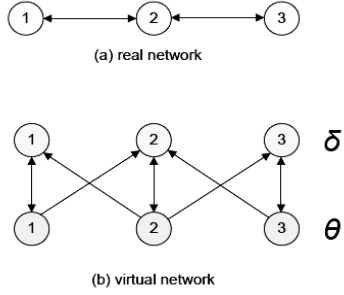

Fig. 1. An illustration of the virtual network with first order dynamics [1].

where $h$ is a negative constant. This means that the generator will linearly increases its mechanical power with rate $h$, proportionally to the frequency drop, with higher order error, when the generator operates close to the stable point $\delta_0$ [1].

### B. Communication based Synchronized Model

By considering a communication network available for the power grid synchronization, except for two physically adjacent generators [1], the second order equation in (8) is converted to a first order one by setting $\theta = \dot{\delta}$ as

$$\begin{cases} \dot{\xi}_i = \theta_i \\ \dot{\theta}_i = \dfrac{1}{M} h \sum_{j \triangleright i}(\xi_i - x_j) - \dfrac{D}{M}\theta + \sum_{i \sim k} \dfrac{V^2 X}{|Z|^2}(\xi_i - \xi_k) \end{cases} \quad (10)$$

where $j \triangleright i$ in the communication network means the generators $i$ and $j$ are adjacent; $h\sum_{j \triangleright i}(\xi_i - x_j)$ denotes the control at generator $i$, where the information form adjacent generators in the communication network is employed. In [1], the variables $\theta_i$ and $\delta_i$ are considered as being in different virtual nodes although they actually belong to the same real node. Therefore, the nodes corresponding to $\theta_i$ and $\delta_i$ are called the 'acceleration node' and the 'angle node', respectively. Such a virtual network is denoted by $\tilde{G}$ as in Fig. 1 [1]. The variable $x_i$ is defined as

$$x_i = \begin{cases} \xi_i & i \leq N \\ \theta_i & otherwise \end{cases} \quad (11)$$

---
1. Since, in practice, the information transmission may not catch up with the physical disturbance propagation [1].

To summarize the dynamics in (10) according to [1] we have

$$\dot{x}_i(t) = c_i x_i + \sigma_{ij} \sum_{i \neq j} a_{ij} x_j \quad (12)$$

where for all nodes in the virtual network $\tilde{G}$ we have

$$a_{ij} = \begin{cases} 1 & i \sim j \\ 0 & otherwise \end{cases} \quad (13)$$

and

$$if \begin{cases} x_i \in set\ of\ \xi \rightarrow & \sigma_{ij} = 0;\ c_i = 1 \\ x_i \in set\ of\ \theta \rightarrow & \sigma_{ij} = \psi;\ c_i = -\dfrac{D}{M} \end{cases} \quad (14)$$

where

$$\psi = \begin{cases} \dfrac{V^2 X}{|Z|^2} & j\ in\ \tilde{G}\ corresponds\ to\ another\ generator \\ \dfrac{h}{M} - \dfrac{V^2 XN}{|Z|^2} & j\ in\ \tilde{G}\ is\ the\ angle\ node\ of\ the\ same\ generator \end{cases} \quad (15)$$

The system dynamics with communication infrastructure is rewritten in matrix form as

$$\dot{x} = \hat{F}x \quad (16)$$

where

$$\hat{F} = \begin{pmatrix} 0 & \hat{I} \\ \dfrac{h}{M}\hat{L}_c + \dfrac{V^2 X}{|Z|^2}\hat{L}_p & -\dfrac{D}{M}\hat{I} \end{pmatrix} \quad (17)$$

where $\hat{L}_p$ is the Laplacian matrix of the power network and $\hat{L}_c$ is that of the communication network. After applying unitary transformation, decoupled system state $\tilde{x}$, can be achieved by solving the characteristic equation

$$M\ddot{\tilde{x}}_i + D\dot{\tilde{x}}_i - (h + \dfrac{V^2 XM\lambda_i}{|Z|^2})\tilde{x}_i = 0 \quad \forall i = 1,...,N \quad (18)$$

which is

$$\phi_{1,2} = \frac{-D \pm \sqrt{D^2 + 4M(h + V^2 XM|Z|^{-2}\lambda_i)}}{2M} \quad (19)$$

The condition for synchronization of the power network is that $\phi_{1,2} < 0$. In this case, we can have the following proposition using [1].

*Proposition 1:* The necessary and sufficient condition for the synchronization of the power network model with communication infrastructure in (10) is

$$\lambda_{max} < \frac{-h|Z|^2}{MV^2 X} \quad (20)$$

where $\lambda_{max}$ is the maximum eigenvalue of the matrix

$\frac{h}{M}\hat{L}_c + \frac{V^2 X}{|Z|^2}\hat{L}_p$ [1]. This matrix can be considered as a Laplacian matrix $\hat{L}_{pc}$ of the following weighted graph

$$(\hat{L}_{pc})_{ij} = \begin{cases} -\omega_{ij} & i \neq j \\ \sum_k \omega_{ik} & i = j \end{cases} \quad (21)$$

where

$$\omega_{ij} = \begin{cases} -\dfrac{h}{M} & i \triangleright j \\ \dfrac{V^2 X}{|Z|^2} & i \sim j \end{cases} \quad (22)$$

## IV. OPTIMUM COMMUNICATION NETWORK DESIGN

In this section, we introduce the proposed method for optimum design of communication infrastructure. In this method, we have employed ACS for selecting the communication lines in the network using the corresponding graph model of network in Fig. 2. In the understudy topology, there is a control center and a number of generators each connected to the center with a communication link. The generators can obtain the measurement information of all other generators connected to the center, which means if $j \triangleright i$, if and only if generators $j$ and $i$ are both connected to the control center. In the proposed method, all possible combinations of generators are considered connected to the center. Then, as the ACS moves forward, useless links are omitted and at the final stage the optimum links remain.

By having in mind the brief ACS descriptions in section II, different steps of the proposed algorithm in pseudocode is illustrated in Fig. 3 and described as follow.

*Initialize:* In this step, initial values of the algorithm parameters such as the number of ants, evaporation coefficients, and the network topology are loaded.

*Locate Ants:* Ants are located on the control center node in this stage. We call the ant which has not been blocked in a junction an alive ant. Since each ant can traverse each node once, an ant is blocked in a node when it has no chance of continuing its transition toward another node and has no possible route to move backward.

*Construct Probability:* When an active ant arrives to a node, the probability of moving from node $i$ to node $j$ for ant $k$ is determined based on its cost function as

$$p_{ij}^k = \begin{cases} \dfrac{\tau_{ij}^\alpha \vartheta_{ij}^{-\beta}}{\sum_{h \notin tabu_k} \tau_{ih}\vartheta_{ij}^{-\beta}} & j \notin tabu_k \\ 0 & otherwise \end{cases} \quad (23)$$

where

$$\vartheta_{ij} = \lambda_{max}\{\frac{h}{M}\hat{L}_{c_{ij}} + \frac{V^2 X}{|Z|^2}\hat{L}_{p_{ij}}\} \quad (24)$$

and $\tau_{ij}$ is the direct route pheromone intensity on the edge between nodes $i$ and $j$; $\alpha$ and $\beta$ controls importance of $\tau_{ij}$ and $\vartheta_{ij}$ respectively and are set to 2 [5,6]; $tabu_k$ is set of blocked edges.

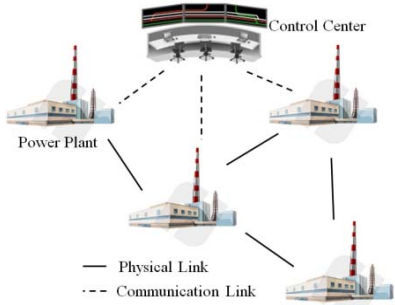

Fig. 2. An illustration of the power network and communication network, which can be modeled to a graph problem.

```
Procedure ACS approach for communication
  network design
  Initialize
  For each iteration
    Locate Ants
    For each ant
      While ant is active
        Construct Probability
        Select Edge
        Update Tabu list
      End While
      Update Local Pheromone
    Next ant
    Update Global Pheromone
  Next iteration
  Select best direction
End ACS approach for communication network
  design
```

Fig.3. Proposed ACS approach for communication network design in pseudocode.

*Select Edge:* A random parameter $0 \leq q \leq 1$ with uniform probability is compared with the parameter $0 \leq Q \leq 1$ which is usually fixed to 0.9 [5, 6]. The comparison result between $Q$ and $q$ picks up one of the two selection methods for the active ant to continue its route to the next node as

$$j = \begin{cases} \arg Max\left(p_{ih}^k\right) & q > Q \\ \text{Roulette Wheel}\left(p_{ih}^k\right) & \text{otherwise} \end{cases}. \quad (25)$$

If $q > Q$, active ant selects edge with the highest probability, otherwise, Roulette Wheel rule is selected to choose the next node through probabilities.

*Update Tabu List:* In this step, the edge that ant $k$ has chosen is added to the tabu list to not be selected again and its probability not be calculated anymore. If ant $k$ is blocked in a node, it is omitted from the active ant list. In other words, this step kills the blocked or arrived ant in the current iteration.

*Update Local Pheromone:* The traditional ACS pheromone system is consisted of two main rules [5] where one is applied whilst constructing solutions (local pheromone update rule) and the other is applied after all ants have finished constructing a solution (global pheromone update rule) [5, 9]. After node selection, the pheromone amount of the edge between nodes $i$ and $j$ is updated for ant $k$ as

$$\tau_{ij}^{new} = \tau_{ij}^{old} + \frac{\gamma}{\vartheta_{ij}} \quad (26)$$

where $\gamma > 1$ is the amount of award. The edge with less cost $\vartheta$ obtains more pheromone award.

*Global Pheromone Update:* The traditional step of each completed loop is global pheromone updating defined as

$$\tau_{ij}^{new} = \rho\, \tau_{ij}^{old} \quad (27)$$

where $0 < \rho < 1$ is evaporation.

*Select Best Direction:* After $m$ iterations, edges with the maximum amount of pheromone are selected as the communication links as

$$n_i^* = \arg Max(\tau_{ij}) \quad \forall i = 1,...,N \quad (28)$$

where $n_i^*$ is list of connected nodes to the node $i$.

## V. NUMERICAL RESULTS

In this section, performance of the proposed ACS approach for communication infrastructure planning in power grids is evaluated. To do so, the 39-

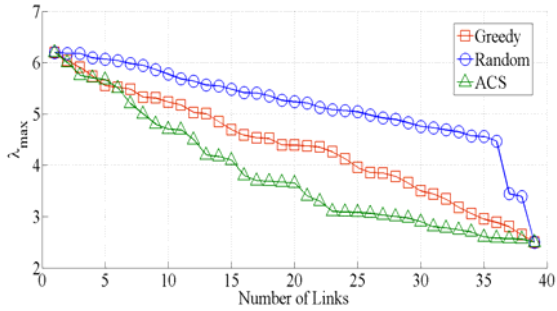

Fig. 4. Comparison of the ACS, greedy exhaustive search, and random communication link selection for different number of links.

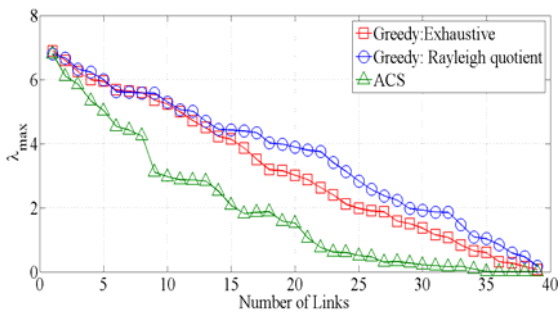

Fig. 5. Comparison of the ACS, greedy exhaustive search in each step, and the greedy Rayleigh quotient approach for different number of links.

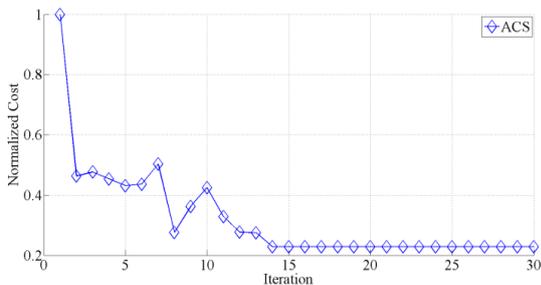

Fig. 6. Normalized cost of the ACS approach in different iterations.

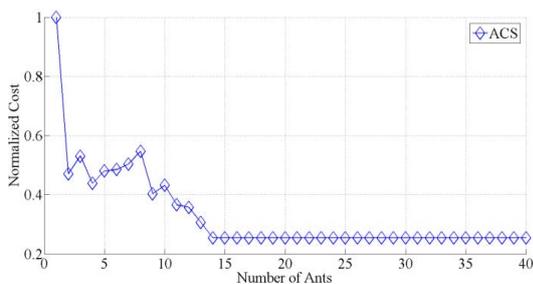

Fig. 7. Normalized cost of the ACS approach for different number of ants.

bus New England power system is used with the assumption that each node has an identical generator [1]. The system is run in 30 iterations with 15 initial ants, $\gamma = 10$, $\rho = 0.9$, $Q = 0.9$, $\alpha = 2$, and $\beta = 2$, where the parameters are set based on trial and error as well as system performance [5,6]. A desktop computer with Intel Core2Quad Q8300 2.5GHz CPU and 3GB of RAM is employed for simulations in MATLAB 2010b environment.

In Fig. 4, performance of the proposed ACS approach is compared with the greedy method in [1] as well as random selection of communication links. Clearly, small largest eigenvalue results in easier synchronization of the system [1]. As this figure illustrates, the random method has the least performance, i.e. has the major largest eigenvalue, among other two methods. This is while the ACS has significant difference from the other two methods and has the best performance. It shows that as the number of links increases and the topology gets more complex, the ACS is still the pioneer among all approaches for different number of links.

In Fig. 5, performance of the greedy algorithms presented in [1], which are the exhaustive search in each step and the Rayleigh quotient based selection, and the proposed ACS approach in this paper are compared. As it is stated in [1], the greedy algorithms have almost very similar behavior in front of different number of links. However, the proposed ACS approach due to its heuristic characteristic as well as search behavior of artificial ants [5, 6], has the smallest eigenvalues for different number of links. This is while the exhaustive search can only guarantee the optimality in each step but cannot guarantee the global optimality [1].

As it is illustrated in Fig. 6, the proposed ACS method has the most normalized cost average in the first cycles of algorithm. As the number of iterations increases, the system arrives to a stable value and by the 14th cycle it is converged. This convergence, which happened not so early-not so late, demonstrates that the parameters of the ACS algorithm are set reasonably [5]. In another evaluation, performance of the ACS approach is evaluated for different number of ants as in Fig. 7. The result demonstrates that by considering one ant in the systems as the link finder, average of normalized costs is so high. However, by activating more ants, average of costs decreases while converging to a specific value of cost average.

## VI. Conclusion and Further Challenges

This paper investigates a new algorithm based on Ant Colony System (ACS) for optimum communication network planning in power grids and develops them toward the smart grids. To do so, the dynamics of power network is first described by 'swing equation'. Since we are using communication infrastructures for the synchronization purpose, the synchronization is determined by weighted sum of the Laplacian matrices of the power network and the communication network together. In order to have an optimum design of the communication network, the above topology is mapped to a graph problem and by employing the ACS algorithm, optimum communication links are selected. The simulation results show performance of the proposed method versus typical approaches in the literature.

In real scenarios, as the scale of network grows, typical algorithms are unable for optimum network planning due to the NP-hard characteristics of problem. However, it has been demonstrated in the literature that the heuristic methods are well-performed tools for such problems. Though, this paper introduces the ACS approach for such problem, several challenges are remained such as incorporating difference of generators into the model and management of infrastructure planning using different techniques such as fuzzy controllers.